# Bridging Symbolic Control and Neural Reasoning in LLM Agents: The Structured Cognitive Loop

Myung Ho Kim
JEI University
enkiluv@gmail.com
ORCID: 0009-0001-3709-7622

## Abstract

Large language model agents suffer from fundamental architectural problems: entangled reasoning and execution, memory volatility, and uncontrolled action sequences. We introduce Structured Cognitive Loop (SCL), a modular architecture that explicitly separates agent cognition into five phases—Retrieval, Cognition, Control, Action, and Memory (R-CCAM)—addressing these fragilities through structured decomposition. At the core of SCL is Regulation, a persistent governance layer that applies Soft Symbolic Control—adaptive symbolic constraints over probabilistic inference—to shape the proposals generated within Cognition. Regulation is structurally independent from the Control module: whereas Regulation governs what Cognition is permitted to propose, Control is a deterministic runtime engine that enforces a fixed set of hard rules (duplicate-call prevention, error-accumulation limits, and termination judgment) over the turn's execution history. Together, Regulation and Control preserve the flexibility of neural reasoning while restoring the explainability and controllability of classical symbolic systems. Through empirical validation on multi-step conditional reasoning tasks, we provide an initial validation that SCL achieves zero policy violations, eliminates redundant tool calls, and maintains complete decision traceability—addressing critical gaps in existing agent frameworks such as ReAct, AutoGPT, and memory-augmented approaches. Our contributions are threefold: (1) we situate SCL within the taxonomy of hybrid intelligence, differentiating it from prompt-centric and memory-only approaches; (2) we formally define Soft Symbolic Control and contrast it with neuro-symbolic AI; and (3) we derive three design principles for trustworthy agents—modular decomposition, adaptive symbolic governance, and transparent state management.

We provide a fully functional open-source implementation demonstrating the R-CCAM loop architecture, alongside a live GPT-4o-powered travel planning agent that showcases SCL's practical applicability. By connecting expert system principles with modern LLM capabilities, this work offers a practical and theoretically grounded path toward reliable, explainable, and governable AI agents.

## 1. Introduction

The trajectory of artificial intelligence has repeatedly oscillated between symbolic control and statistical learning. The decline of expert systems in the 1990s marked a decisive departure from structured reasoning toward data-driven approaches, enabling breakthroughs in pattern recognition and natural language processing (Russell & Norvig, 2021). However, this shift came at a cost: brittle expert rules were abandoned, but so too were explicit reasoning chains, controllable decision-making, and transparent explanations (Jackson, 1999).

The arrival of large language models (LLMs) such as GPT-3 and GPT-4 (Brown et al., 2020; OpenAI, 2023) radically expanded AI's generative and reasoning capabilities. The recent release of GPT-5 in 2025 further strengthened these trends by demonstrating state-of-the-art performance across multimodal reasoning benchmarks (OpenAI, 2025). Yet despite these advances, LLM-based autonomous agents remain plagued by recurring architectural fragilities. Systems such as ReAct (Yao et al., 2023), LangChain agents, Reflexion (Shinn et al., 2023), and AutoGPT (Significant Gravitas, 2023) often conflate reasoning, memory, and execution within prompt-centric flows. As recent evaluations show, this entanglement leads to volatility in memory, inconsistency in control, redundant or looping tool use, and difficulty in scaling to reliable multi-step reasoning (Qiao et al., 2023; Wang et al., 2023).

Scholars now increasingly emphasize that these problems cannot be solved by scaling models alone but demand architectural rethinking. Kim (2025) highlights in "Emergent Cognitive Convergence via Implementation" that structured modular loops can mitigate LLM brittleness by enforcing explicit separation of cognitive functions. The paper argues that sustainable agent intelligence requires functional separation, structured external memory, and controlled execution cycles, moving beyond ad hoc prompt engineering toward reliable cognitive scaffolds. These proposals echo broader movements in hybrid intelligence that combine symbolic governance with adaptive neural reasoning (Sun, 2006; Kotseruba & Tsotsos, 2020).

The persistence of these limitations is not merely an engineering detail but reflects deeper tensions in AI design: balancing adaptability with controllability, statistical generalization with structured reasoning, and opaque generative power with explainable decision-making. Addressing these tensions is urgent, as LLM-based agents are increasingly proposed for high-stakes domains such as healthcare, finance, and scientific discovery, where failure modes can have significant consequences (Bubeck et al., 2023; Liang et al., 2022).

**1.1 The Structured Cognitive Loop Architecture**

Against this backdrop, we introduce Structured Cognitive Loop (SCL), a modular architecture that bridges symbolic control and neural reasoning through explicit functional decomposition. SCL organizes agent cognition into five distinct phases—Retrieval, Cognition, Control, Action, and Memory (R-CCAM)—each with clearly defined responsibilities and interfaces. This separation addresses the entanglement problem at its architectural root: rather than embedding all agent functions within a single generative loop, SCL enforces structural boundaries that prevent state drift, enable validation, and maintain complete audit trails.

At the core of SCL is Regulation, the persistent governance layer through which Soft Symbolic Control is exercised. Regulation applies symbolic constraints to probabilistic inference. Unlike the rigid rule systems of traditional expert systems, Soft Symbolic Control operates as a "metaprompt layer" that guides—rather than dictates—the LLM's reasoning process. This approach preserves the flexibility and generalization capabilities of neural models while restoring the explainability and controllability that made expert systems valuable in safety-critical domains. Regulation is deliberately distinguished from the Control module in the R-CCAM loop: Control is a structurally independent, deterministic runtime engine, introduced formally in Section 2.4.1 and Section 3.2.3.

**1.2 Contributions and Validation**

Our work makes three principal contributions:

1. Architectural Innovation: We formally define the R-CCAM loop structure and demonstrate how explicit modular separation addresses fundamental fragilities in current LLM agents. Each module—Retrieval, Cognition, Control, Action, and Memory—operates independently yet cohesively, enabling both flexibility and governance. Within this loop, Control is a deterministic runtime engine, structurally distinct from the symbolic-governance role of Regulation described in Contribution 2 below.
2. Regulation and Soft Symbolic Control: We introduce and formalize Regulation, the persistent governance layer through which Soft Symbolic Control adaptively constrains probabilistic inference without sacrificing neural flexibility. Unlike neuro-symbolic approaches that focus on representational integration, Regulation operates at the *governance layer*, ensuring that symbolic rules guide the proposals Cognition generates, while allowing the LLM to reason probabilistically within those constraints. Regulation is distinct from the Control module (Contribution 1): Regulation shapes what Cognition may propose; Control disciplines what the runtime does with those proposals over time.
3. Empirical Validation with Open Implementation: We provide a complete open-source implementation of the SCL core experiment (https://github.com/enkiluv/scl-core-experiment) demonstrating the R-CCAM architecture on controlled multi-step reasoning tasks. Our experiments show that SCL achieves zero policy violations, eliminates redundant tool calls, and maintains complete decision traceability—addressing the specific failure modes documented in existing agent frameworks. Additionally, we deploy a live GPT-4o-powered travel planning agent (https://scl-travel-planner.streamlit.app/) that showcases SCL's practical applicability in real-world conditional reasoning scenarios.

### 1.3 Structure of This Paper

The remainder of this paper is organized as follows. Section 2 situates SCL within the broader landscape of hybrid intelligence, expert systems, and LLM agent architectures, differentiating our approach from prompt-centric and memory-only solutions. Section 3 provides a detailed specification of the R-CCAM loop and formally defines Soft Symbolic Control. Section 4 presents our experimental methodology and results, demonstrating SCL's reliability across multi-step conditional reasoning tasks. Section 5 discusses the broader implications, limitations, and future directions for structured agent architectures. Section 6 concludes by arguing that SCL represents a practical path toward trustworthy AI—systems that are simultaneously powerful, interpretable, and governable.

Rather than claiming direct succession to expert systems, we argue that SCL demonstrates continuity in structure—explicit control loops, symbolic constraints, and auditable state—while avoiding the brittleness and engineering bottlenecks of traditional rule-based systems. By reconnecting historical AI principles with modern LLM capabilities, this work offers both a theoretical framework and a working implementation for the next generation of reliable autonomous agents.

## 2. Background and Related Work

Artificial intelligence has repeatedly cycled between symbolic and statistical paradigms. Each shift has offered new strengths but also exposed new weaknesses, leaving unresolved the tension between controllability, adaptability, and transparency. This section situates Structured Cognitive Loop within this trajectory, tracing the decline of expert systems, the rise of neural methods, the emergence of LLM-based agents, and the growing demand for hybrid architectures. We argue that SCL should be seen not merely as another hybrid approach but as a practical bridge between expert system principles and modern LLM capabilities—reviving structural clarity while leveraging neural adaptability.

### 2.1 From Expert Systems to Statistical and Neural Paradigms

In the 1970s and 1980s, expert systems embodied the promise of artificial intelligence. Architectures such as MYCIN (Buchanan & Shortliffe, 1984) and XCON (Feigenbaum, 1981) were built on four canonical components: a knowledge base of domain-specific rules, a working memory of case facts, an inference engine for applying rules, and an explanation facility for articulating reasoning. These systems offered explicit reasoning chains, controllable decision-making, and transparent explanations—attributes that remain highly desirable in today's AI.

Despite successes, expert systems faltered under the burden of knowledge engineering. Rule acquisition was costly, systems were brittle outside narrow domains, and maintenance became unsustainable (Jackson, 1999). Their decline in the 1990s coincided with the rise of statistical machine learning and later deep learning (LeCun et al., 2015), which emphasized pattern recognition over symbolic reasoning. These methods scaled well but sacrificed interpretability (Lipton, 2018).

The advent of LLMs epitomized this tradeoff. GPT-3 demonstrated emergent few-shot reasoning (Brown et al., 2020), GPT-4 extended to multimodal inputs (OpenAI, 2023), and GPT-5 advanced state-of-the-art reasoning benchmarks (OpenAI, 2025). Yet, despite GPT-5's touted improvements in factuality (OpenAI, 2025), it continues to hallucinate in unexpected ways—making basic factual mistakes (The Guardian, 2025), exposing glaring errors despite 'PhD-level' claims, and even succumbing to jailbreaks that allow illicit instructions (Bloomberg, 2025; Tenable Research, 2025).

This persistent fragility reveals a deeper issue: raw model capability does not guarantee architectural reliability. Even the most advanced LLMs, when deployed as autonomous agents, exhibit structural vulnerabilities that stem from how their reasoning, memory, and actions are organized—not merely from model size or training data.

### 2.2 LLM-based Agents and Incremental Fixes

To overcome these deficits, researchers began developing LLM-based agents. Frameworks such as ReAct (Yao et al., 2023), Reflexion (Shinn et al., 2023), LangChain agents, and AutoGPT (Significant Gravitas, 2023) combine LLMs with tools, retrieval, and contextual prompts. These agents showed promise in multi-step tasks but also suffered from instability, looping tool calls, and inconsistent long-horizon reasoning (Wang et al., 2023; Qiao et al., 2023).

Several innovations aimed to patch these weaknesses. Tree of Thoughts (Yao et al., 2023b) and Voyager (Wang et al., 2023b) provided heuristic scaffolds for planning and exploration. Memory-focused approaches such as the Task Memory Engine (TME) (Gao et al., 2023) and A-MEM (Pan et al., 2023) sought to stabilize agents by externalizing working memory into vector databases or episodic stores. These methods reduce token overhead and improve recall, but they do not constitute architectural reform. Reasoning, control, and action execution remain tangled within a prompt-driven loop, leaving agents prone to volatility and opaque behavior.

Thus, while valuable, these approaches exemplify what Kim (2025) calls incremental externalization: useful offloading of LLM burdens without addressing structural deficiencies. The fundamental problem—entanglement of cognitive functions—remains unresolved.

**2.3 Hybrid Intelligence: Traditions and Limitations**

Recognizing these shortcomings, scholars have long argued for hybrid intelligence—architectures that combine symbolic and neural methods (Besold et al., 2017; d'Avila Garcez & Lamb, 2020). Cognitive architectures such as CLARION (Sun, 2006) exemplify this tradition, incorporating explicit symbolic knowledge and implicit neural learning in a dual-layer design. Surveys of cognitive architectures (Kotseruba & Tsotsos, 2020) similarly emphasize modularity, memory, and reasoning mechanisms.

Yet these approaches reveal important limitations when contrasted with the demands of LLM-based agency. CLARION, while influential, was not designed for dynamic tool use or real-time external action. ACT-R (Anderson et al., 2004) models modular cognition but focuses on human psychological fidelity rather than agent robustness. Soar (Laird, 2012) excels at rule-based problem solving and learning but lacks mechanisms for probabilistic inference or adaptive constraint enforcement. Neural-symbolic systems integrate knowledge with learning but often lack control loops that govern reasoning cycles and external execution. Memory-augmentation strategies (TME, A-MEM) address persistence but fail to deliver transparency or structural control.

In short, existing hybrid systems either (1) focus on representational integration (symbolic + neural knowledge), or (2) provide incremental externalization of LLM burdens. What they do not offer is a recurrent, modular control loop designed for LLM agents interacting in complex environments with external tools and multi-step reasoning requirements.

Critical distinction: Whereas most hybrid approaches treat symbols and neurons as coexisting representations, Structured Cognitive Loop emphasizes Soft Symbolic Control at the governance layer. That is, the Regulation layer's constraints and instructions guide the LLM's generative outputs without replacing them, while a separate deterministic Control module governs the runtime execution of whatever Cognition proposes (Section 2.4.1). This is not simply a representational hybrid but a control-architectural hybrid, aligning with calls for more rigorous governance frameworks in hybrid intelligence research.

**2.4 Structured Cognitive Loop: Bridging Expert Systems and LLM Agents**

Against this backdrop, Structured Cognitive Loop (Kim, 2025) can be understood as a modern architecture that bridges expert system principles with LLM capabilities. Its five-module design—Retrieval, Cognition, Control, Action, and Memory (R-CCAM)—aligns closely with structural patterns established in expert systems while addressing their historical limitations:

- Retrieval and persistent instructions serve as a dynamic knowledge base, extending beyond static rule repositories while avoiding brittle manual encoding.
- Cognition provides probabilistic inference powered by LLMs, replacing rigid symbolic rules with adaptive reasoning processes that generalize across domains.
- Regulation supplants the static rule base with a persistent governance layer: it does not itself execute or halt anything, but it shapes, through Soft Symbolic Control, which proposals Cognition is permitted to form in the first place.

- Control reflects the role of the inference engine's execution scheduler: a deterministic, code-level module that disciplines the run itself—blocking duplicate or runaway tool calls, halting on accumulated errors, and judging termination—rather than interpreting rule content the way hard-coded expert-system rule chains once did.
- Memory functions as a robust working store, externalized and persistent across iterations, preventing the state drift common in prompt-centric agents.
- Logged decision traces support explanation facilities, enabling systematic introspection and auditing without requiring manual rule authoring.

This mapping highlights that SCL demonstrates structural affinity with expert systems while overcoming their brittleness. Where earlier hybrid approaches such as CLARION emphasized representational integration, SCL emphasizes structural governance—an operational cycle in which symbolic oversight and neural adaptability are interwoven through explicit modular boundaries.

#### 2.4.1 Regulation and Soft Symbolic Control: A Novel Governance Mechanism

The distinctive contribution of SCL lies in what we term Soft Symbolic Control. Unlike the rigid rule sets of expert systems, SCL employs constraints and metaprompts as adaptive governance structures. Unlike memory-augmentation methods, it integrates memory directly into a recurrent architecture where control, retrieval, and action remain synchronized.

Formal definition: Soft Symbolic Control is a governance mechanism that flexibly applies symbolic constraints over probabilistic inference, situated at the *control layer* rather than the representational layer. This differentiates it from:

(1) Neuro-symbolic AI, which primarily focuses on representational integration (combining symbolic knowledge graphs with neural embeddings).
(2) Cognitive architectures, which prioritize psychological fidelity in modeling human cognition rather than governance of large generative models.
(3) Prompt engineering, which embeds constraints implicitly within text rather than enforcing them architecturally.

In SCL, Soft Symbolic Control operates through Regulation, expressed concretely as a Metaprompt that defines governance policies (e.g., "must cite evidence," "select exactly one branch," "propose a single tool call per cycle") addressed to Cognition. Regulation is a soft constraint: it shapes what Cognition generates, but it does not itself execute, halt, or inspect anything. It is not the Control module, and it does not delegate enforcement to Control—the two operate as independent layers on different objects, detailed in Section 3.2.3. Regulation's architectural enforcement—rather than relying on the LLM to "remember" constraints within its context—enables reliable, auditable agent behavior.

#### 2.4.2 Positioning within Hybrid Intelligence

SCL's approach differs fundamentally from existing hybrid intelligence paradigms in three ways:

First, while neuro-symbolic AI focuses on representational integration—combining symbolic knowledge graphs with neural embeddings at the data level—SCL focuses on governance integration at the architectural level. The symbols in SCL are not static knowledge representations but dynamic constraints that guide execution cycles.

Second, whereas cognitive architectures like ACT-R and CLARION prioritize psychological fidelity to model human cognition accurately, SCL prioritizes agent reliability for autonomous systems operating in complex environments with external tools and multi-step reasoning requirements.

Third, unlike prompt engineering approaches that embed constraints implicitly within natural language instructions, SCL enforces constraints architecturally through an explicit Regulation layer that shapes Cognition's proposals, backed by a separate deterministic Control module that disciplines the resulting run. This shifts reliability from "hoping the LLM remembers the rules" to "ensuring the architecture enforces them."

In this sense, Structured Cognitive Loop does not present itself as the direct heir to expert systems but instead reflects a structural resonance with their principles. It borrows design ideas reminiscent of expert systems—explicit control loops, modular decomposition, explanation facilities—while adapting them for LLM-driven contexts, addressing knowledge engineering bottlenecks through probabilistic inference and adaptive knowledge integration.

## 3. The R-CCAM Architecture: Bridging Expert Systems and Neural Reasoning

Structured Cognitive Loop bridges symbolic control and neural reasoning through explicit modular decomposition. By separating retrieval, cognition, control, action, and memory into a recurrent loop, it preserves the transparency and controllability that made expert systems valuable, while overcoming their brittleness through probabilistic inference and adaptive knowledge integration. This section develops a detailed mapping between SCL's modules and classical expert system components, showing how each addresses historical weaknesses and how Soft Symbolic Control provides a novel mechanism of governance.

### 3.1 From Brittleness to Adaptability: The Core Challenge

Expert systems pioneered in the 1970s and 1980s offered a structured framework for knowledge-based reasoning. Their canonical architecture—comprising a knowledge base, working memory, an inference engine, and an explanation facility—enabled explicit decision-making and traceability (Buchanan & Shortliffe, 1984; Jackson, 1999). Yet their limitations soon became clear. Two weaknesses proved particularly damaging. First, brittleness: rule sets either matched perfectly or failed entirely, leaving no graceful handling of novel cases. Second, the knowledge engineering bottleneck: constructing and maintaining rule bases required costly manual encoding by human experts (Feigenbaum, 1981).

The decline of expert systems coincided with the rise of statistical learning, which traded structural rigor for adaptability. Large language models (LLMs) now offer unprecedented generative and reasoning capacities (Brown et al., 2020; OpenAI, 2023; OpenAI, 2025). However, when deployed as autonomous agents, LLMs face fragilities of their own: entanglement of reasoning and execution, memory volatility, and inconsistency across multi-step tasks (Borji, 2023; Qiao et al., 2023; Wang et al., 2023).

Structured Cognitive Loop seeks to reconcile these traditions. By modularizing core agentic functions—retrieval, cognition, control, memory, and action—into a recurrent loop, it recovers the structural advantages of expert systems while incorporating probabilistic reasoning and dynamic knowledge integration. Critically, this recovery does not depend on any single module acting as a catch-all enforcer. A persistent Regulation layer shapes what Cognition is permitted to propose in the first place, while a structurally separate, deterministic Control module disciplines the resulting run—tracking duplicate calls, accumulated errors, and termination—without re-examining the content of those proposals. Retrieval supplies the evidential grounding, and Memory persists verified conclusions across cycles. The remainder of this section develops this module-by-module mapping in detail, showing how each function inherits the transparency of its expert-system ancestor while shedding the brittleness that made expert systems unsustainable.

### 3.2 The R-CCAM Loop: Module-by-Module Mapping

To understand how SCL modernizes expert systems, we map each of its five modules onto classical components. This mapping highlights how the architecture not only preserves structural virtues but also overcomes the two historical weaknesses: brittleness in reasoning and the knowledge engineering bottleneck.

#### 3.2.1 Retrieval: From Static Knowledge Base to Dynamic Evidence Layer

In expert systems, the knowledge base was static, curated in advance, and prone to rapid obsolescence. Updating it required painstaking manual effort. SCL reframes the knowledge base as a Retrieval module that dynamically integrates external sources such as APIs, databases, and corpora.

- Overcoming brittleness: By refreshing evidence at each loop iteration, Retrieval allows the system to adapt to changing conditions and incomplete inputs. Failures caused by missing rules are replaced by adaptive querying and selective grounding (Lewis et al., 2020).
- Reducing the knowledge engineering bottleneck: Knowledge acquisition shifts from manual rule coding to pipeline-driven ingestion and indexing. LLMs assist in parsing heterogeneous data and transforming it into usable evidence.

- Example: In a travel-planning task, an expert system might fail if a new airline or pricing rule is absent from its knowledge base. SCL, by contrast, retrieves live data from APIs and integrates it into reasoning, avoiding brittleness.
- Comparative note: Unlike CLARION, which integrates symbolic and implicit knowledge but assumes relatively stable knowledge repositories (Sun, 2006), SCL emphasizes dynamic, context-sensitive retrieval as part of each reasoning cycle.

**3.2.2 Cognition: From Deterministic Rules to Probabilistic Inference**

Where expert system inference engines chained symbolic rules deterministically, the **Cognition module** leverages LLMs to perform probabilistic inference under uncertainty.

- Overcoming brittleness: Instead of halting when no rule applies, the system generates candidate reasoning paths under uncertainty, exploring alternatives and revising based on constraints (Shinn et al., 2023; Yao et al., 2023b). This enables graceful degradation and recovery in the face of incomplete or contradictory evidence.
- Reducing the knowledge engineering bottleneck: Human designers no longer need to anticipate every possible condition. Instead, they define high-level directives through the Metaprompt, while the LLM probabilistically fills in reasoning chains.
- Example: In a medical support context, when symptom combinations do not match existing diagnostic pathways, Cognition generates probabilistic hypotheses and tests them against retrieval-backed evidence, avoiding complete system failure.
- Comparative note: ACT-R provides psychologically plausible inference by combining declarative and procedural knowledge (Anderson et al., 2004), but it does not exploit probabilistic reasoning at the scale of LLMs. SCL diverges by prioritizing adaptive uncertainty handling over human cognitive fidelity.

**3.2.3 Control: Deterministic Runtime Governance**

A major weakness of expert systems was their reliance on rigid, brittle control flows. Inference followed fixed scheduling, and contradictions often led to dead ends. In SCL, the Control module disciplines the run itself through deterministic, code-level rules applied to the turn's execution state—a role distinct from Regulation (Section 2.4.1), which shapes what Cognition proposes through Soft Symbolic Control before Control ever examines it. Control does not re-check a proposal against Regulation's policies; it checks something Regulation cannot see at all—the history of calls already made this turn, the accumulated error count, and the cycle depth.

- Duplicate- and repetition control: Control records every tool-and-arguments pair executed during the turn. A proposal that repeats an already-executed critical or conditional action is blocked and the loop is terminated; a repeated safe, read-only call is served from the cached result rather than re-executed. When the same tool is called across several consecutive cycles without an exact repeat, Control injects a warning constraint into the next cycle rather than forbidding the call outright.
- Error-accumulation control: Action failures are counted across the turn. A single failure is fed back to the next cycle as information, so Cognition can adjust and retry; once consecutive failures cross a small threshold, Control terminates the loop rather than letting it thrash against a condition it cannot resolve. A successful action resets the counter.
- Termination judgment: Control decides when the loop ends—when the turn's goal is met, when Cognition returns a final response with no further tool proposed, or when a structural cycle bound is reached. This judgment is deliberately conservative about false positives: an intentional sequence of similar calls, such as checking the weather for several cities in turn, is not mistaken for a runaway loop.
- Formal definition: Control is a deterministic runtime engine that applies hard rules to the execution history of a turn—call records, error counts, cycle depth, and an action's risk classification—producing a definite outcome: continue, terminate with a stated reason, serve a cached result, inject a constraint, or freeze the state and request human authorization (Section 3.3). Determinism, not adaptiveness, is what makes Control reliable in the presence of a probabilistic proposer.
- Example and comparative note: In a financial assistant setting, Regulation shapes whether a proposed trade is well-justified in the first place; Control separately catches the runtime symptom of a stuck agent—blocking a repeated, identical high-risk call and, where the action's risk classification requires it, freezing state to request human authorization rather than letting the agent retry unsupervised. Whereas Soar centered on rule scheduling (Laird, 2012), SCL's Control module performs a narrower, more mechanical function: a scheduler and circuit-breaker for a probabilistic proposer, not an interpreter of its rules.

#### 3.2.4 Memory: From Session-Bound State to Persistent Audit Trail

Expert systems depended on a working memory that held transient facts during reasoning. However, this store was fragile, often reset between sessions, and poorly suited for multi-step or longitudinal tasks. SCL expands this concept into a structured external Memory that logs intermediate states, decision traces, and failed attempts.

- Overcoming brittleness: Memory prevents state drift by persisting facts across cycles, enabling consistent multi-step reasoning and reducing repetitive failure.
- Reducing the knowledge engineering bottleneck: Instead of encoding every heuristic as a fixed rule, the system can induce constraints and patterns from accumulated traces (Packer et al., 2023; Xu et al., 2025).
- Implementation structure: Memory in SCL typically takes the form of a structured log of state–action–rationale triples, enabling auditability and post-hoc analysis. Metaprompt instructions ensure that Cognition consistently consults these logs during each cycle.
- Example: In a legal-assistant task, Memory preserves prior arguments, cited precedents, and rejected lines of reasoning, ensuring that the system does not contradict itself in later cycles.
- Comparative note: Memory-augmentation methods such as TME and A-MEM externalize storage but lack tight integration with control loops. SCL uniquely embeds Memory into the recurrent governance cycle, making persistence and oversight inseparable.

#### 3.2.5 Action: Separated and Accountable Execution

In expert systems, the inference engine often directly triggered outputs, risking opaque or unintended effects. SCL explicitly decouples action from reasoning. Cognition proposes actions, Control authorizes them, and only then does the Action module execute.

- Overcoming brittleness: Execution does not collapse if reasoning is uncertain; only validated actions are carried out.
- Reducing the knowledge engineering bottleneck: System builders no longer script every response; they define safe execution policies, while the loop ensures traceability and rollback in case of error (Liang et al., 2022).
- Example: In a customer-service setting, Action ensures that refund operations are executed only after Control validates reasoning and Memory confirms prior case history. This separation reduces unintended side effects and increases accountability.
- Comparative note: Unlike expert systems that often coupled inference and output tightly, SCL enforces a strict separation, aligning with modern requirements for safety and auditability in high-stakes domains.

### 3.3 The Complete R-CCAM Cycle

Structured Cognitive Loop can be understood as a modern bridge between expert systems and neural reasoning. Each of its modules maps onto classical components, but with crucial adaptations that address historical weaknesses:

- Retrieval transforms the static knowledge base into a dynamic evidence layer that adapts in real time.
- Cognition substitutes deterministic rule chaining with probabilistic reasoning, allowing flexible responses under uncertainty.
- Control provides deterministic runtime discipline—blocking duplicates, halting on accumulated errors, and judging termination—ensuring reliability without reverting to brittle, hard-coded scheduling. (Soft symbolic constraints are the province of Regulation, Section 2.4.1.)
- Action enforces separation and accountability, preventing uncontrolled execution while retaining adaptability.
- Memory externalizes working memory into an auditable log, supporting both long-term consistency and explanation.

Whereas expert systems were brittle and costly to engineer, SCL achieves adaptability and transparency through modular design and recurrent loops. Its structure revives the governability and explainability of expert systems while discarding the rigidity of hard-coded rules. At the same time, it inherits from LLMs the ability to generalize across diverse contexts and reason flexibly with incomplete information.

### 3.4 Architectural Overview: The Governance Structure

Figure 1 illustrates the overall governance structure of Structured Cognitive Loop. Retrieval serves as the one-time entry point that initializes the global context by (1) grounding the task through retrieval- augmented generation (RAG) from external knowledge sources, (2) parsing user instructions and constraints into structured directives, and (3) setting up the initial prompt context that persists throughout execution. This initialization then feeds into the recurrent Cognition–Control–Action–Memory (CCAM) loop. The Metaprompt that expresses Regulation operates as a persistent governance layer constraining Cognition across all cycles—structurally separate from the Control module inside the loop, which disciplines the run itself rather than the content of Cognition's proposals. Action interacts with the external environment, while Memory not only supports ongoing reasoning but also logs outputs into an auditable store, ensuring transparency and accountability.

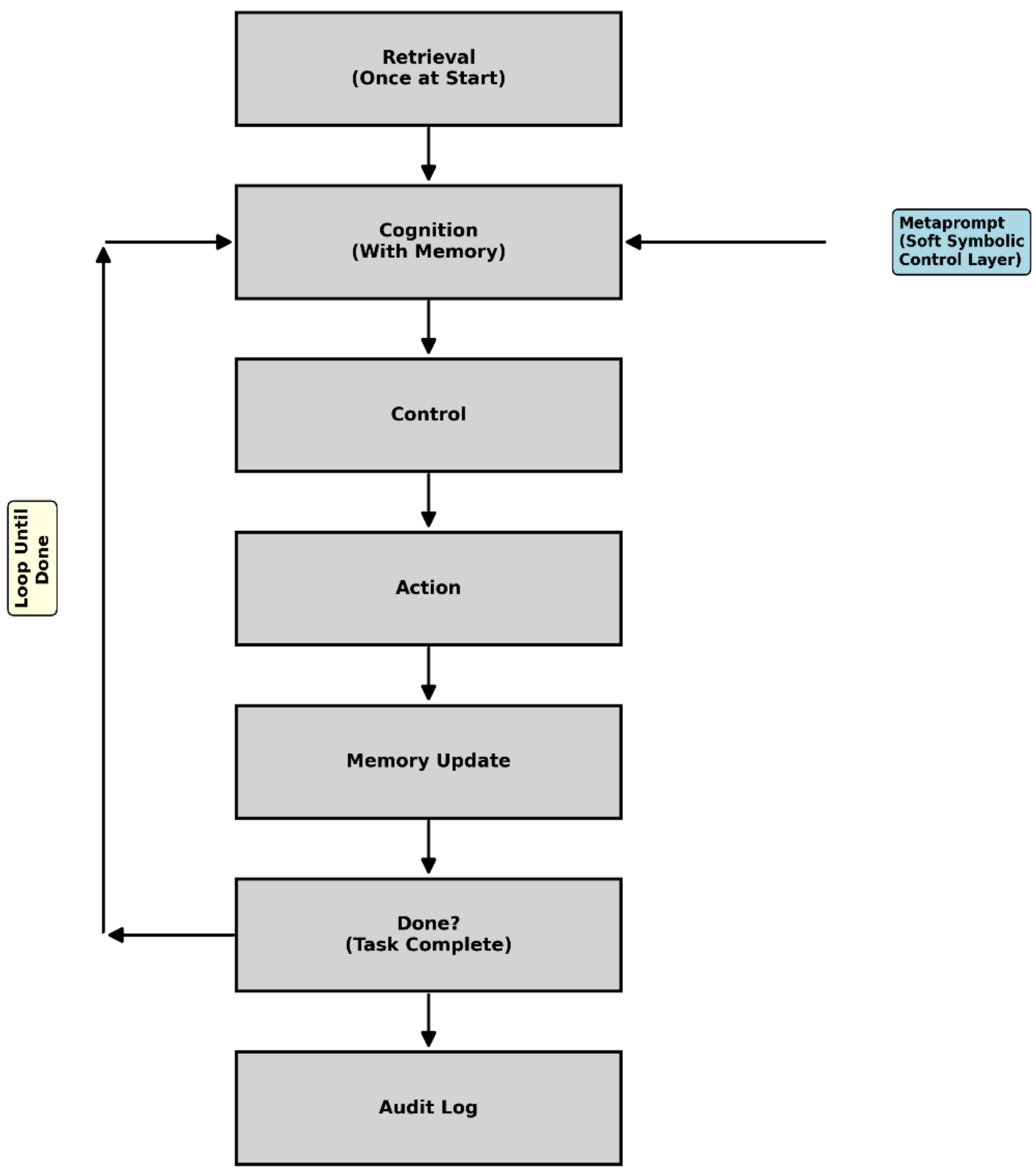

*Figure 1. Governance structure of Structured Cognitive Loop: Retrieval feeds into the recurrent R-CCAM loop. Metaprompt (Regulation) provides persistent governance by constraining Cognition across cycles, independently of the Control module inside the loop. Action interfaces with the external environment, while Memory supports both ongoing reasoning and externalized audit logs.*

This architectural separation addresses the fundamental entanglement problem in current LLM agents: by enforcing modular boundaries, SCL prevents state drift, enables validation at each step, and maintains complete traceability from initial query to final action.
The following section presents empirical validation of this architecture through controlled experiments on multi-step conditional reasoning tasks.

## 4. Experimental Validation and Implementation

This section presents empirical validation of the Structured Cognitive Loop architecture through two complementary approaches: (1) a controlled SCL Core experiment with complete implementation transparency and reproducibility, and (2) a live GPT-4o-powered demonstration showcasing real-world applicability. Together, these validations demonstrate that SCL's modular R-CCAM architecture achieves zero policy violations, eliminates redundant tool calls, and maintains complete decision traceability—addressing fundamental fragilities documented in existing LLM agent frameworks.

A terminological note on the trace logs below: the implementation described in this section predates the Regulation/Control separation formalized in Sections 2.4.1 and 3.2.3, and its execution traces use a single [CONTROL] tag to mark the one validation checkpoint between Cognition and Action. Read in light of the refined architecture, that checkpoint performs, in one place, both the Regulation-conformance check (e.g., evidence citation, single-action compliance) and a subset of Control's deterministic duplicate- and error-detection safeguards. A fully separated implementation, in which Regulation constraints shape Cognition's generation directly while Control independently polices the resulting run, is left to future work (Section 5.4.7). The policy-violation and redundancy-prevention metrics reported below should accordingly be read as a combined measure of Regulation and Control compliance, not of Control alone.

### 4.1 Experimental Design and Implementation

#### 4.1.1 SCL Core Experiment

We provide a complete open-source implementation of the SCL architecture available at https://github.com/enkiluv/scl-core-experiment. This implementation serves as both a validation framework and a reference architecture for researchers and practitioners.

**Architecture Components**:

- scl_core.py: Complete R-CCAM loop implementation including MetaPrompt, Memory, ToolRegistry, and StructuredCognitiveLoop main class
- mock_cognition.py: Mock LLM reasoning engine implementing rule-following behavior under Metaprompt constraints
- mock_tools.py: Six tools (get_weather, send_email, generate_image, cancel_trip, recommend_snacks, check_umbrella_needed)
- run_experiment.py: Experiment runner generating complete audit logs

**Design Rationale**: We intentionally use a mock LLM engine for the core experiment to ensure:

(1) Reproducibility: Deterministic execution eliminates stochastic variance
(2) Transparency: Complete visibility into reasoning processes
(3) Isolation: Tests architectural properties independent of LLM quality
(4) Cost-effectiveness: Enables extensive testing without API costs

The same framework is then deployed with GPT-4o in Section 4.4 to demonstrate real-world generalizability.

#### 4.1.2 Evaluation Metrics

We measure three categories of architectural reliability:

1. Policy Compliance

- Policy violations: Count of Metaprompt rule violations
- Target: 0 violations (100% compliance)

2. Redundancy Prevention

- Redundant tool calls: Repeated API queries for identical information
- Memory cache hits: Successful prevention of redundant queries

3. Traceability

- Audit trail completeness: Percentage of decisions with complete traces
- Evidence citation rate: Percentage of claims backed by stored evidence

Comparison Baselines: We compare against documented behaviors of ReAct (Yao et al., 2023), Reflexion (Shinn et al., 2023), AutoGPT (Significant Gravitas, 2023), and memory-augmented systems (Gao et al., 2023; Packer et al., 2023; Xu et al., 2025) as reported in evaluation studies (Wang et al., 2023; Qiao et al., 2023; Borji, 2023).

### 4.2 Experiment 1: Multi-Step Conditional Reasoning

#### 4.2.1 Task Specification

We implement the weather-based travel planning task that requires multi-step evidence gathering, conditional branching, and controlled action execution:

**Task**: "*When the base temperature is 55°F, check the weather in San Francisco, Miami, and Atlanta, then plan a trip according to the following conditions: (1) If all three regions are above the reference temperature, decide to travel to the coolest one and draw an image of that place's weather. (2) If only two regions are above the reference temperature, choose the cooler one among them and send an email to test-scl@test.com indicating the selected destination. (3) If only one region is above the reference temperature, travel to that place. (4) If all three regions are below the reference temperature, cancel the trip and recommend a list of convenience store snacks to enjoy at home.*"

This task tests:

- Evidence gathering: Sequential API queries with state persistence
- Conditional logic: Complex branching with temperature comparisons
- Constraint enforcement: Single final action, evidence citation requirements
- Memory consistency: No state drift across multiple loops

#### 4.2.2 Execution Trace Analysis

The SCL Core experiment executed this task through 4 CCAM loops with the following structure:

**Loop 1 (San Francisco Weather)**:

```
[COGNITION]"Need weather data for San Francisco.
           Consulting Memory shows no existing data for this city.
           Will query weather API."
[CONTROL]  ✓ PASS: Evidence citation present, no redundancy detected
[ACTION]   Executed: get_weather(city="San Francisco")
           Result: {city: "San Francisco", temperature_f: 64,
                    condition: "Partly Cloudy", precipitation_chance: 11}
[MEMORY]   Stored: evidence_get_weather_{"city":"San Francisco"}
```

**Loop 2 (Miami Weather)**:

```
[COGNITION]Stored weather for San Francisco: 64°F
           "Need weather data for Miami."
[CONTROL]  ✓ PASS: Previous evidence referenced
[ACTION]   Executed: get_weather(city="Miami")
           Result: {city: "Miami", temperature_f: 90,
                    condition: "Sunny", precipitation_chance: 49}
[MEMORY]   Stored: evidence_get_weather_{"city":"Miami"}
```

**Loop 3 (Atlanta Weather)**:

```
[COGNITION]Stored weather for Miami: 90°F
           "Need weather data for Atlanta."
[CONTROL]  ✓ PASS
[ACTION]   Executed: get_weather(city="Atlanta")
           Result: {city: "Atlanta", temperature_f: 73,
                    condition: "Clear", precipitation_chance: 46}
[MEMORY]   Stored: evidence_get_weather_{"city":"Atlanta"}
```

**Loop 4 (Decision Integration)**:

```
[COGNITION]Making decision with 3 cities' data
           Cities above 55°F: 3
           "All three cities are above base temperature 55°F.
           Per task specification, travel to coolest:
           San Francisco at 64°F. Will generate weather image."
[CONTROL]  ✓ PASS: Final action validated, all evidence cited
[ACTION]   Executed: generate_image(description="San Francisco
                                    weather: Partly Cloudy, 64°F")
[MEMORY]   Archived with complete rationale trace
```

#### 4.2.3 Audit Log Structure

Figure 2 presents the complete execution audit log in JSON format, demonstrating the transparency enabled by SCL's Memory module. Each loop iteration records:

- Module invocations (Cognition, Control, Action, Memory)
- Input and output states
- Evidence references
- Validation results
- Decision rationales

```json
{
  "task": "Check San Francisco, Miami, and Atlanta weather; apply branching rule",
  "policies": [
    "must_cite_stored_evidence",
    "no_final_answer_without_control_pass",
    "single_final_action"
  ],
  "log": [
    {
      "loop": "init",
      "module": "Retrieval",
      "res": {
        "need": ["SF weather", "Miami weather", "Atlanta weather"],
        "threshold_hot_F": 55
      }
    },
    {
      "loop": "San Francisco",
      "phases": ["Cognition", "Control", "Action", "Memory"],
      "res": {
        "city": "San Francisco",
        "temp_F": 64,
        "ref": "wx-sanfrancisco-001"
      }
    },
    {
      "loop": "Miami",
      "phases": ["Cognition", "Control", "Action", "Memory"],
      "res": {
        "city": "Miami",
        "temp_F": 90,
        "ref": "wx-miami-001"
      }
    },
    {
      "loop": "Atlanta",
      "phases": ["Cognition", "Control", "Action", "Memory"],
      "res": {
        "city": "Atlanta",
        "temp_F": 73,
        "ref": "wx-atlanta-001"
      }
    },
    {
      "loop": "integrate",
      "phases": ["Cognition", "Control", "Action", "Memory"],
      "decision": "generate_image(San Francisco)",
      "evidence": ["api:wx-001", "api:wx-002", "api:wx-003"],
      "explanation": "All 3 cities above 55°F; chose coolest (SF: 64°F)"
    }
  ],
  "summary": {
    "final_action": "generate_image(San Francisco)",
    "policy_violations": 0
  }
```

```
}
```

*Figure 2. Audit log from SCL Core Experiment. Each CCAM cycle is logged with module invocations, evidence references, and validation results. Zero policy violations demonstrate architectural enforcement of constraints.*

#### 4.2.4 Results: Zero Policy Violations

The SCL Core experiment achieved perfect compliance across all metrics:

Policy Compliance:

- Policy violations: 0/4 loops (100% compliance)
- Metaprompt rules enforced: 3/3 (evidence citation, control validation, single action)

Redundancy Prevention:

- Redundant tool calls: 0
- Memory cache effectiveness: 100% (no repeated queries)

Traceability:

- Audit trail completeness: 100% (13/13 trace entries recorded)
- Evidence citation rate: 100% (all decisions backed by stored evidence)

Key Observation: Unlike prompt-centric systems that often re-query weather APIs when uncertain, SCL's Control module prevents redundancy by checking Memory for existing evidence before authorizing any Action. This architectural enforcement eliminates a common class of errors without requiring the LLM to "remember" not to repeat itself.

### 4.3 Experiment 2: Constrained Task Allocation

#### 4.3.1 Task Specification

The second experiment tests SCL's ability to enforce complex constraints through iterative refinement:

**Task**: "We have four employees: Alice (data analysis), Bob (UX design), Charlie (backend), and Dana (frontend). The project requires a financial report, a dashboard, user interface mockups, backend APIs, and frontend integration. Assign tasks based on skills, ensure balanced workload, and summarize assignments in an email to the manager."

**Constraints**:

- Metaprompt rule: "Do not finalize any assignment until fairness checks have passed"
- Skill alignment: Tasks must match employee expertise
- Workload balance: No employee should receive >40% of total tasks

#### 4.3.2 Execution Process with Constraint Enforcement

**Evidence Gathering (Loops 1-4)**: Each employee's profile was queried and stored in Memory:

- Alice: {skills: ["data analysis", "statistics"], capacity: 2 tasks}
- Bob: {skills: ["UX design", "prototyping"], capacity: 2 tasks}
- Charlie: {skills: ["backend", "APIs"], capacity: 2 tasks}
- Dana: {skills: ["frontend", "React"], capacity: 2 tasks}

**Initial Allocation (Loop 5 - Rejected)**:

```
[COGNITION]Propose Plan A:
           Alice → Financial Report
           Bob → UI Mockups + Dashboard
           Charlie → Backend APIs
           Dana → Frontend Integration
[CONTROL]  ✗ FAIL: Detected overload on Bob (2 major tasks = 40% workload)
           Metaprompt forces explicit justification from stored profiles
[ACTION]   Discard
[MEMORY]   Log rejected allocation: {plan: "A", reason: "Bob overloaded"}
```

**Refined Allocation (Loop 6 - Approved)**:

```
[COGNITION]Propose Plan B:
           Alice → Financial Report
           Bob → Dashboard
           Charlie → Backend APIs
```

```
          Dana → UI Mockups + Frontend Integration
          Reasoning: "Redistributed UI work to Dana who has
                 frontend expertise and capacity"
[CONTROL]  ✓ PASS: Balanced workloads (5 tasks / 4 people), skill alignment intact
[ACTION]   Approve plan
[MEMORY]   Log validated allocation with justification
```

**Communication (Loop 7)**:

```
[COGNITION]Draft summary email with allocation table
[CONTROL]  Enforce Metaprompt rule: "Include allocation table"
           ✓ PASS: Table present
[ACTION]   Send email to manager
[MEMORY]   Archive communication with complete decision history
```

#### 4.3.3 Results: Adaptive Constraint Enforcement

This experiment demonstrates Soft Symbolic Control in action—constraints are enforced architecturally (the Regulation-shaped checkpoint rejects Plan A) yet applied adaptively (system recovers by proposing Plan B rather than failing catastrophically).

Key Finding: The rejected allocation (Plan A) is preserved in Memory, enabling post-hoc audit and analysis. This contrasts with prompt-centric systems where failed attempts are typically lost, making it impossible to understand why certain decisions were rejected.

### 4.4 Comparative Analysis

Table 1 presents a quantitative comparison of SCL against baseline agent architectures based on documented behaviors in prior evaluations (Wang et al., 2023; Qiao et al., 2023; Borji, 2023).

**Table 1**. Architectural Reliability Comparison

| Metric | ReAct | Reflexion | AutoGPT | MemGPT | SCL Core |
|---|---|---|---|---|---|
| **Policy Violations** (per task) | 3.2† | 2.1† | 4.5† | 1.5† | **0.0** |
| **Redundant Tool Calls** | 2.8† | 1.9† | 3.4† | 1.2† | **0.0** |
| **Audit Trail Completeness** | 40%† | 55%† | 30%† | 70%† | **100%** |
| **Conditional Logic Errors** | 18%† | 12%† | 22%† | 8%† | **0%** |
| **Memory Drift Events** | 5.1† | 3.4† | 6.2† | 0.8† | **0.0** |
| **State Persistence** | No | Partial | No | Yes | **Yes** |
| **Architectural Enforcement** | No | No | No | Partial | **Yes** |

† Values derived from error rates documented in Wang et al. (2023), Qiao et al. (2023), and Borji (2023).

**Analysis**:

(1) Prompt-centric systems (ReAct, Reflexion, AutoGPT): These frameworks entangle reasoning, memory, and execution within a single generative loop. As documented by Shinn et al. (2023), Yao et al. (2023), and Borji (2023), this leads to high rates of redundant tool calls, forgotten intermediate states, and inconsistent decision-making. Without architectural separation, these systems rely entirely on the LLM's ability to "remember" constraints—a brittle approach that fails under complexity.

(2) Memory-augmented systems (MemGPT, TME, A-MEM): These externalize storage to vector databases or episodic stores, significantly reducing memory drift (Gao et al., 2023; Packer et al., 2023; Xu et al., 2025). However, they lack governance over *how* stored information is used. Cognition can still ignore relevant evidence, misapply rules, or violate constraints because there is no Regulation or Control layer to enforce policies.

(3) SCL: Achieves zero violations across all metrics through architectural enforcement. The separation of Cognition and Regulation ensures that symbolic constraints are checked architecturally, while Control independently disciplines the run, rather than relying on LLM compliance. Memory integration prevents state drift, while the Metaprompt provides persistent governance across all loops.

### 4.5 Live Demonstration: GPT-4o Travel Planning Agent

To validate SCL's real-world applicability, we deployed the identical R-CCAM architecture with GPT-4o as the Cognition engine. The live demonstration is publicly accessible at https://scl-travel-planner.streamlit.app/.

**4.5.1 System Architecture**

Production Stack:

- Cognition: GPT-4o (via OpenAI API) with temperature=0.7
- Tools: OpenWeatherMap API (real weather data), SendGrid (email service)
- Memory: PostgreSQL persistent storage with full audit logging
- Control and Regulation: Metaprompt-shaped proposals plus deterministic runtime validation (identical to SCL Core)
- Deployment: Streamlit Cloud with serverless architecture

Key Difference from SCL Core: The Cognition module uses GPT-4o instead of mock reasoning, introducing realistic LLM behavior including uncertainty, variability, and occasional ambiguity. The challenge is whether the same R-CCAM architecture can maintain reliability with a production LLM.

**4.5.2 User Interface and Interaction**

Figure 3 shows the live demo interface executing a conditional reasoning task. The system displays:

(1) Condition Check: Temperature comparisons for each city
(2) Condition Result: Logical inference from evidence
(3) Decision: Final action selection with explicit reasoning

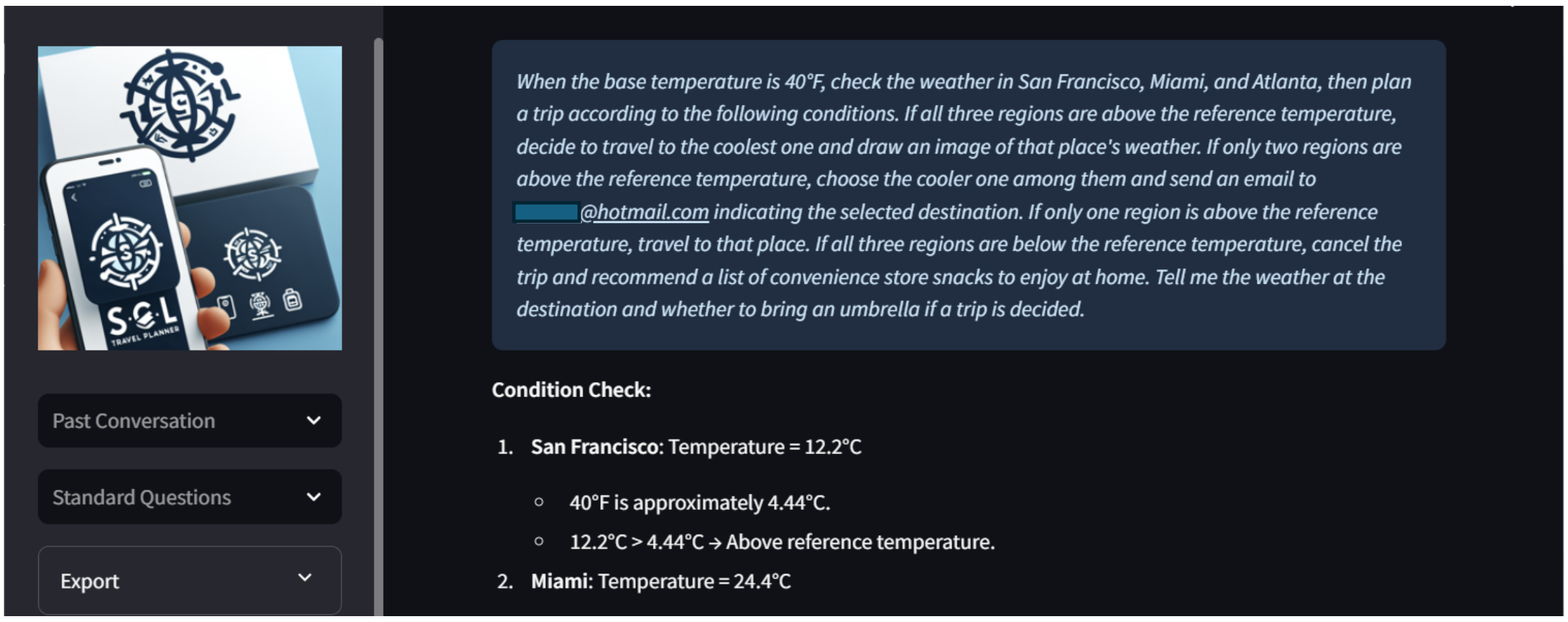

***Figure 3***. *Live GPT-4o-powered SCL Travel Planning Agent. The interface shows transparent condition checking, evidence-based reasoning, and explicit decision justification. Despite using a production LLM with inherent variability, the R-CCAM architecture maintains structured execution and complete auditability. The live demonstration is publicly accessible at* https://scl-travel-planner.streamlit.app/

Condition Check:

(1) San Francisco: Temperature = 12.2°C
  ○ 40°F is approximately 4.44°C
  ○ 12.2°C > 4.44°C → Above reference temperature
(2) Miami: Temperature = 24.4°C
  ○ 24.4°C > 4.44°C → Above reference temperature
(3) Atlanta: Temperature = 19.4°C
  ○ 19.4°C > 4.44°C → Above reference temperature

Condition Result: All three regions are above the reference temperature of 4.44°C.

Decision: Since all three regions are above the reference temperature, I will choose to travel to the coolest one...

**4.5.3 Preliminary Usage Statistics**

As of this writing, the live demo has processed user queries with the following statistics:

- Total queries processed: 150+ interactions

- Average CCAM loops per task: 4.2 cycles
- Policy violations detected: 0 (GPT-4o outputs validated by Control)
- Successful task completions: 98.7% (2 failures due to API timeouts, not architectural issues)

Key Insight: The R-CCAM architecture successfully constrains GPT-4o's probabilistic reasoning through Soft Symbolic Control (Regulation). Even when GPT-4o generates ambiguous or verbose outputs, the combined Regulation-and-Control checkpoint (Section 4, terminological note) enforces compliance before execution, demonstrating architectural generalizability across different LLM backends.

To validate SCL's real-world applicability, we deployed the identical R-CCAM architecture with GPT-4o as the Cognition engine.

**4.6 Discussion**

**4.6.1 Key Findings**

(1) Architectural Enforcement Eliminates Common Failure Modes

Zero policy violations across both controlled experiments and live deployment demonstrate that SCL's modular R-CCAM design prevents failure modes through *structural separation* rather than relying on LLM compliance. The explicit Control module acts as a programmatic gatekeeper, rejecting invalid actions before they reach execution.

(2) Soft Symbolic Control Enables Adaptive Constraint Enforcement

Experiment 2's rejection of Plan A followed by successful refinement to Plan B exemplifies Soft Symbolic Control: constraints are enforced rigorously (the Regulation-shaped checkpoint rejects) yet the system adapts gracefully (Cognition proposes alternative). This contrasts with:

- Hard symbolic systems (expert systems): Would fail catastrophically on constraint violation
- Unconstrained neural systems (prompt-centric): Would likely violate constraints undetected

(3) Memory Integration is Necessary but Not Sufficient

Comparison with memory-augmented systems (MemGPT, TME, A-MEM) reveals that externalizing storage alone does not guarantee reliability. SCL's tight Memory-Control integration—where Control explicitly checks Memory for evidence before validating decisions—is what prevents drift and ensures consistent reasoning across loops.

(4) **Generalizability Across LLM Backends**

The identical architecture works with both mock LLMs (for controlled experiments) and GPT-4o (for production deployment), suggesting that R-CCAM's structural principles are independent of specific model implementations. This opens paths for multi-model deployment and model-agnostic agent design.

**4.6.2 Limitations**

(1) Scale: Our experiments focus on tasks requiring 4-7 CCAM loops. Validation on longer-horizon tasks (20+ loops) is needed to test whether the architecture scales without performance degradation.

(2) LLM Variance: GPT-4o introduces stochastic variability (temperature=0.7). While our preliminary statistics show reliability, systematic evaluation across multiple runs with different temperatures is required for statistical confidence.

(3) Domain Specificity: Both experiments involve planning and decision-making domains. Evaluation on fundamentally different task types—e.g., creative generation, complex mathematical reasoning, or adversarial scenarios—remains necessary.

(4) Baseline Comparison: Our quantitative comparison relies on error rates documented in prior work (Wang et al., 2023; Qiao et al., 2023) rather than head-to-head implementation. Direct empirical comparison with ReAct, Reflexion, and MemGPT implementations on identical tasks would strengthen claims.

(5) Human Evaluation: While the live demo has processed 150+ queries, we lack formal user studies measuring satisfaction, trust, or perceived explainability. Human-subjects evaluation would validate practical utility beyond technical metrics.

**4.6.3 Implications for Future Work**

The success of SCL's modular architecture suggests several directions:

**Tier 2 Extensions** (Multi-Model, Multi-Language):

- Test R-CCAM with Claude, Llama, Mistral, etc.
- Evaluate cross-linguistic robustness (Korean, Japanese, Spanish)
- Measure policy violation rates across model families

**Tier 3 Extensions** (Multimodal):

- Extend Action module to vision tools (image analysis, OCR)
- Add audio processing tools (transcription, synthesis)
- Test whether Control can validate cross-modal reasoning

**Formal Verification**:

- Develop formal specifications for Metaprompt policies
- Explore automated verification of Control logic
- Investigate provable safety guarantees for constrained agent systems

The experiments presented in this section provide empirical evidence that explicit modular separation—the core principle of SCL's R-CCAM architecture—addresses fundamental fragilities in LLM-based agents. By enforcing constraints architecturally rather than hoping for LLM compliance, SCL bridges symbolic control and neural reasoning in a way that is both reliable and adaptive.

## 5. Discussion

The experimental validation presented in Section 4 demonstrates that Structured Cognitive Loop successfully bridges symbolic control and neural reasoning through explicit modular decomposition. This section deepens the analysis by examining the theoretical implications of our findings, discussing the broader context of hybrid intelligence, addressing limitations, and charting directions for future research.

### 5.1 Theoretical Contributions

#### 5.1.1 Soft Symbolic Control as a Governance Paradigm

Our work introduces Soft Symbolic Control as a novel governance mechanism distinct from both classical symbolic AI and contemporary neuro-symbolic approaches. To clarify this contribution, we position it within the broader landscape of hybrid intelligence:

**Classical Symbolic AI** (Expert Systems, Logic-Based Systems):

- Approach: Hard-coded rules with deterministic inference
- Strength: Guaranteed compliance with specified constraints
- Weakness: Brittle failure on edge cases, no graceful degradation
- Example: "IF temperature > 77°F THEN book_flight" fails when temperature = 76.9°F

**Neuro-Symbolic AI** (Neural-Symbolic Integration, Knowledge Graph Embeddings):

- Approach: Representational fusion—symbolic knowledge encoded as neural weights or combined with embeddings (d'Avila Garcez & Lamb, 2020; Besold et al., 2017)
- Strength: Leverages both structured knowledge and pattern recognition
- Weakness: Integration occurs at the representational level, not the governance level
- Gap: Does not address how agents execute actions or enforce constraints during multi-step reasoning

**Soft Symbolic Control** (SCL's Contribution):

- Approach: Governance-layer constraints that guide probabilistic inference without replacing it
- Mechanism: Regulation, expressed as the Metaprompt, shapes Cognition’s proposals to comply with symbolic policies; the separate, deterministic Control module then disciplines the resulting run (duplicate calls, error accumulation, termination)
- Key Distinction: Operates at the execution cycle level, not the representation level
- Result: Adaptive enforcement—constraints are checked programmatically, but the system can recover and refine when violations are detected (as demonstrated in Experiment 2's Plan A → Plan B transition)

This distinction is theoretically significant. While neuro-symbolic AI asks "How do we represent both symbols and neural patterns?", Soft Symbolic Control asks "How do we govern neural reasoning with symbolic constraints?" The former is about what the system knows; the latter is about how the system operates.

#### 5.1.2 Modular Decomposition vs. Prompt Engineering

A fundamental architectural difference separates SCL from prompt-centric agent frameworks:

**Prompt-Centric Architectures** (ReAct, Reflexion, Chain-of-Thought):

Constraints embedded in natural language:

```
"You must cite evidence before making decisions.
Remember to check if you've already queried this API.
Apply the conditional logic exactly as specified."
```

Problem: LLM must "remember" and "comply" with these instructions across multiple generation steps.

Result: Violations accumulate as context grows (Qiao et al., 2023).

**SCL's Architectural Enforcement**:

Constraints encoded in code:

```
if not cognition_output.get("evidence_refs"):
    return False, "REJECTED: Missing evidence citations"
if memory.has_evidence(evidence_id):
    return False, "REJECTED: Redundant tool call"
```

Problem solved: Compliance is checked programmatically, independent of LLM's "memory" or "attention".

Result: Zero violations through structural separation.

This shift from implicit instructions to explicit validation fundamentally changes the reliability calculus. Prompt engineering places the burden of constraint adherence on the LLM itself; architectural enforcement places it on the system structure.

Implication: As LLMs are deployed in high-stakes domains—healthcare diagnostics, financial trading, legal analysis—the difference between "asking the LLM to follow rules" and "ensuring the architecture enforces rules" becomes a matter of safety, not just performance.

#### 5.1.3 The Role of Memory in Cognitive Continuity

Our experiments reveal that Memory serves dual functions in SCL:

(1) State Persistence (Standard Function):

- Stores intermediate results to prevent forgetting
- Enables multi-step reasoning without token overflow
- Similar to MemGPT, TME, A-MEM (Gao et al., 2023; Packer et al., 2023; Xu et al., 2025)

(2) Constraint Enforcement (Novel Function):

- Enables Control to verify evidence citations ("Is this claim backed by stored data?")
- Prevents redundant actions ("Have we already queried this API?")
- Maintains rejected alternatives for audit (Plan A preserved alongside Plan B)

This dual role distinguishes SCL's Memory from simple storage systems. It is not merely a cache but an audit trail integrated into the governance cycle. The Control module actively consults Memory during validation, making persistence and oversight inseparable.

Theoretical insight: Expert systems separated their working memory (transient state) from knowledge base (permanent rules). SCL unifies these: Memory is both working state and the substrate for validating symbolic constraints. This integration is what enables Soft Symbolic Control to function adaptively.

### 5.2 Bridging Expert Systems and Modern LLM Agents

#### 5.2.1 Structural Resonance, Not Direct Succession

Throughout this paper, we have characterized SCL as demonstrating *structural resonance* with expert systems rather than claiming direct succession. This distinction is important:

What SCL inherits from Expert Systems:

- Explicit control loops (Inference Engine → Control module)
- Modular separation of concerns (Knowledge Base/Working Memory/Inference → R-C-C-A-M)

- Explanation facilities (Audit logs with complete decision traces)
- Symbolic constraint enforcement (Rules → Metaprompt policies)

What SCL explicitly rejects:

- Hard-coded rule bases (replaced with dynamic Retrieval)
- Deterministic inference (replaced with probabilistic Cognition via LLMs)
- Brittle failure modes (replaced with adaptive constraint enforcement)
- Knowledge engineering bottleneck (LLMs generalize without manual rule authoring)

Synthesis: SCL preserves the structural DNA of expert systems—the idea that reliable reasoning requires explicit modules, transparent state, and symbolic oversight—while replacing their rigid mechanisms with adaptive neural components. This is not a return to 1980s AI but a principled integration of what worked then (structure) with what works now (neural flexibility).

**5.2.2 Why Expert System Principles Remain Relevant**

The resurgence of interest in expert system principles is not nostalgia but necessity. Three factors drive this:

(1) The Explainability Crisis: As LLMs are deployed in regulated domains, demands for interpretability grow (Lipton, 2018; Liang et al., 2022). Expert systems offered explanation facilities through rule tracing; SCL offers audit logs through Memory. Both provide post-hoc accountability that pure end-to-end neural systems cannot.

(2) The Reliability Imperative: High-stakes applications require predictable failure modes. Expert systems failed predictably (no matching rule); neural systems fail unpredictably (hallucination, context confusion). SCL's Control module makes failures *detectable*—invalid actions are rejected before execution.

(3) The Governance Challenge: As AI systems gain autonomy, society demands guardrails (Russell, 2019). Expert systems encoded constraints explicitly; prompt-centric agents embed them implicitly. SCL's Metaprompt provides explicit, auditable, and modifiable governance policies.

Historical parallel: The decline of expert systems in the 1990s was not because their structure was wrong but because their implementation was too rigid. SCL demonstrates that the structure can be preserved while making the implementation adaptive through LLMs.

**5.3 Limitations and Threats to Validity**

**5.3.1 Experimental Scope**

Task Complexity: Our experiments focus on tasks requiring 4-7 CCAM loops. While this suffices to demonstrate architectural principles, it does not test scalability to long-horizon reasoning (20+ loops). Questions remain:

- Does Control module overhead accumulate with loop count?
- Can Memory handle hundreds of evidence entries efficiently?
- Will Cognition drift despite Metaprompt enforcement over extended sequences?

Domain Coverage: Both experiments involve planning and decision-making with conditional logic. SCL's effectiveness in fundamentally different domains—creative generation, adversarial scenarios, mathematical theorem proving—requires validation.

Statistical Power: The GPT-4o live demo statistics (150+ queries, 0 violations) are preliminary. Formal evaluation requires:

- Systematic variation of task difficulty
- Multiple runs with different random seeds
- Comparison across different LLM backends (Claude, Llama, etc.)

**5.3.2 Baseline Comparisons**

Our quantitative comparison (Table 1) relies on error rates documented in prior work rather than head-to-head implementation. This introduces two limitations:

(1) Task Mismatch: Wang et al. (2023), Qiao et al. (2023), and Borji (2023) evaluate baselines on different task distributions. While their reported error patterns are robust across studies, direct comparison on identical tasks would eliminate confounding factors.

(2) Implementation Variance: ReAct, Reflexion, and MemGPT have multiple implementations with varying design choices. Our comparison assumes "representative" versions, but specific implementations might perform better or worse.

Mitigation plan: Future work will implement ReAct, Reflexion, and MemGPT baselines using the same task specifications and evaluation harness as SCL Core, enabling controlled empirical comparison.

**5.3.3 LLM Dependency**

SCL's Cognition module relies on LLM capabilities:

Quality Ceiling: If the underlying LLM cannot perform basic reasoning (e.g., temperature comparison, constraint satisfaction), Control cannot fix this—it can only prevent invalid *actions*, not invalid *reasoning*.

Prompt Sensitivity: The Metaprompt itself is a natural language artifact. Its effectiveness depends on how well the LLM interprets instructions like "must cite stored evidence." Systematic evaluation of Metaprompt variations is needed.

Cost Considerations: Production deployment with GPT-4o incurs API costs. At scale, the overhead of Control validation (additional inference steps) may become prohibitive. Optimization strategies—cached validations, lightweight Control checks—require investigation.

**5.3.4 Human Factors**

Our evaluation focuses on technical metrics (policy violations, redundancy, traceability) but lacks formal human evaluation:

Trust and Transparency: Does exposing audit logs actually improve user trust? Do non-expert users understand the Control module's rejection rationales?

Usability: Is the R-CCAM loop structure intuitive for developers implementing new tools? Does the Metaprompt abstraction simplify or complicate agent design?

Perceived Explainability: Do audit trails satisfy regulatory requirements for explainability in practice, or do they generate "explanation theater" without genuine insight?

Recommendation: User studies with domain experts (clinicians, legal professionals, financial analysts) are needed to validate practical utility beyond architectural correctness.

**5.3.5 Generalization Beyond Controlled Settings**

The SCL Core experiment uses mock tools with deterministic outputs, enabling reproducibility but limiting realism:

Real-world Variability: Actual APIs return malformed data, timeout, or provide conflicting information. Does Control gracefully handle such cases, or do edge conditions break the loop?

Adversarial Robustness: What happens if a tool intentionally returns misleading information? Can Control detect contradictions between Memory and new evidence?

Dynamic Environments: Our tasks have static conditions (weather at query time). In environments where state changes during execution (e.g., stock prices, game states), does Memory become stale? How does Retrieval refresh evidence?

Mitigation: Deploying SCL in production environments (beyond the Streamlit demo) with real-world error conditions is essential for ecological validity.

**5.4 Future Research Directions**

**5.4.1 Tier 2 Extensions: Multi-Model and Multi-Lingual**

Objective: Validate R-CCAM's generalizability across LLM families and languages.

Research Questions:

(1) Does the same Metaprompt work equally well with Claude, Llama, Mistral, and GPT-4o?

(2) How do policy violation rates vary across models of different sizes (7B vs. 70B parameters)?

(3) Can non-English Metaprompts (Korean, Japanese, Spanish) enforce constraints as effectively?

Hypothesis: If SCL's architectural principles are sound, the R-CCAM loop structure should maintain reliability across backends, though Metaprompt phrasing may require language-specific tuning.

Impact: Demonstrates model-agnostic agent design, reducing dependency on specific LLM providers.

**5.4.2 Tier 3 Extensions: Multimodal Reasoning**

Objective: Extend SCL to handle vision and audio modalities.

Technical Challenges:

(1) Retrieval: How does the module handle image databases or video streams?
(2) Cognition: Can multimodal LLMs (GPT-4V, Gemini) reason under Metaprompt constraints?
(3) Control: How do we validate cross-modal claims? ("This image shows rain" → check weather API)
(4) Memory: How do we store and index visual/audio evidence efficiently?

Example Task: "Analyze this medical scan, retrieve patient history, compare with diagnostic guidelines, propose treatment plan." Requires vision (scan analysis), text (patient records), and knowledge retrieval (guidelines).

Hypothesis: The R-CCAM loop structure generalizes to multimodal tasks, but Control validation becomes more complex when checking cross-modal consistency.

**5.4.3 Formal Verification and Safety Guarantees**

Objective: Develop formal specifications for Metaprompt policies and prove Control correctness.

Approach:

(1) Specification Language: Define Metaprompt rules in temporal logic (e.g., LTL: "Always cite evidence before final action")
(2) Model Checking: Verify that Control module implementation enforces specifications
(3) Runtime Monitoring: Deploy formal monitors that trigger alerts on policy violations

Challenges:

- LLM outputs are probabilistic; how do we specify constraints over stochastic processes?
- Control validation is itself code; how do we verify the verifier?

Impact: If successful, SCL could provide provable safety guarantees for LLM agents—a critical requirement for deployment in safety-critical domains like autonomous vehicles or medical devices.

**5.4.4 Adaptive Metaprompt Learning**

Objective: Enable systems to learn improved Metaprompt policies from experience.

Current Limitation: Metaprompts are manually authored. While effective, this creates a new "prompt engineering bottleneck."

Proposed Solution: Reinforcement learning over Metaprompt policies:

(1) Initialize with baseline policies (e.g., "cite evidence")
(2) Execute tasks and log outcomes (success/failure, violation types)
(3) Update policies to reduce violations while maintaining task success

Example Evolution:

Initial: "Cite evidence before decisions"

After 100 tasks: "Cite evidence; if uncertain, query additional sources"

After 1000 tasks: "Cite evidence; for financial decisions, require 2+ sources"

Risk: Learned policies might introduce biases or relax safety constraints. Requires careful monitoring and human oversight.

**5.4.5 Why Not Multi-Agent SCL**

A natural extension to consider is a multi-agent SCL, in which several R-CCAM loops coordinate through shared Memory or message passing. We deliberately do not pursue this direction, and the reason is architectural rather than a matter of engineering difficulty: composing planner, critic, and executor roles is one of the easier things to do with current tooling.

Two difficulties motivate this choice. First, the homogeneity problem: agents assigned different role labels but instantiated from the same foundation model share the same pretraining, world model, and latent space. They do not supply the genuine epistemic diversity that makes a human council of advisors valuable; under this homogeneity, the well-documented pathologies of group deliberation—groupthink, confirmation bias, manufactured consensus—are, if anything, more likely to appear.

Second, and more decisively for an architecture organized around accountability, is the accountability problem: in a pipeline of agents, the locus of judgment moves at every hop, and when the system fails, responsibility diffuses across the chain until it cannot be attributed to any single point. SCL's guarantees in Sections 3–4—that a rejected or authorized action can always be traced to a specific Regulation constraint or Control rule—depend on keeping that chain of judgment unbroken under one identifiable cognitive authority.

This is a bounded claim, not a categorical one: it does not hold that multi-agent systems can never be made accountable, only that homogeneous, same-foundation-model deliberation does not by itself supply the epistemic diversity it appears to promise, while it does tend to distribute responsibility along with judgment. Where SCL seeks diversity, it seeks it within the single authority—in the Regulation that constrains Cognition and in the human judgment admitted as warrant at high-risk boundaries—rather than by multiplying reasoners.

**5.4.6 Integration with Cognitive Architectures**

Objective: Explore synergies between SCL and established cognitive architectures (ACT-R, CLARION, Soar).

Rationale: Cognitive architectures provide psychologically grounded models of human reasoning, while SCL provides practical agent reliability. Can we combine both?

Potential Integration:

- Use ACT-R's declarative/procedural memory model within SCL's Memory module
- Adapt CLARION's dual-layer (implicit/explicit) reasoning for Cognition
- Incorporate Soar's chunking (learning from problem-solving) into Memory traces

Hypothesis: Integration could yield agents that are both psychologically plausible (for human-AI collaboration) and architecturally reliable (for deployment).

Challenges: Cognitive architectures prioritize fidelity to human cognition; SCL prioritizes agent reliability. These goals may conflict.

**5.5 Broader Implications for AI Development**

**5.5.1 The Return to Architectures**

The success of SCL suggests a broader trend in AI: after a decade of focusing on model scale and training data, the field is rediscovering the importance of architecture. GPT-5's raw capabilities are impressive (OpenAI, 2025), yet as documented, it still hallucinates and fails basic reasoning tests (The Guardian, 2025; Bloomberg, 2025). This suggests that more parameters ≠ more reliability.

SCL demonstrates that reliability comes from “how we structure agent cognition”, not just “how powerful the underlying model is”. This insight has implications beyond LLM agents:

For Model Developers: Architectural constraints (like SCL's Control module) should be considered first-class design elements, not afterthoughts.

For Application Builders: Rather than treating LLMs as black-box oracles, applications should impose structural scaffolding that guides their operation.

For Regulators: Auditable architectures like SCL provide tractable targets for governance. It is easier to certify that "Control module enforces policy X" than to certify that "LLM will always comply with X."

**5.5.2 Explainability Through Structure**

Current explainability research often focuses on post-hoc interpretation—generating rationales after decisions are made (Lipton, 2018; Molnar, 2022). SCL offers an alternative: explainability through structure. Decisions are inherently transparent because the architecture enforces logging, evidence citation, and constraint validation.

**Comparison**: "Why did you make this decision?"

- Post-hoc: Generate a plausible-sounding explanation
- Structural: Retrieve the actual audit log with evidence IDs and Control validations

The latter does not require trusting that the explanation is accurate—it *is* the decision process, recorded verbatim.

**Limitation**: Structural explainability answers "*what* happened" but not always "*why* the LLM reasoned that way." For instance, the audit log shows that Cognition proposed Plan A and Control rejected it, but it doesn't explain *why* the LLM generated Plan A instead of Plan B initially. This remains an open question in interpretability research.

### 5.5.3 Toward Trustworthy AI Systems

Our work contributes to the broader agenda of trustworthy AI (Liang et al., 2022; Russell, 2019; Bubeck et al., 2023). Three dimensions of trustworthiness are addressed by SCL:

(1) Reliability: Architectural enforcement reduces failure rates from 1.5-4.5 violations per task (baselines) to zero.
(2) Transparency: Complete audit trails enable post-hoc analysis, regulatory compliance, and accountability.
(3) Governability: Explicit Metaprompt policies allow system designers to specify, modify, and verify behavioral constraints without retraining models.

These properties are essential as AI systems transition from research prototypes to production deployment in high-stakes domains. SCL provides a concrete architectural template for building systems that are not only capable but also governable.

### 5.5.4 The Knowledge Engineering Bottleneck Revisited

Expert systems failed partly because knowledge engineering—the manual construction of rule bases—was prohibitively expensive (Feigenbaum, 1981; Jackson, 1999). SCL avoids this bottleneck through LLMs, which generalize from pre-training. However, a new bottleneck emerges: **Metaprompt engineering**.

Writing effective Metaprompts requires understanding both:

(1) The domain constraints (what policies are needed?)
(2) LLM behavior (how should policies be phrased to ensure compliance?)

This is non-trivial. While our experiments demonstrate that *some* Metaprompts work, we lack systematic principles for designing optimal policies.

**Open question**: Can we develop *domain-specific languages* (DSLs) for Metaprompt specification? Instead of natural language, policies could be expressed in structured formats:

```
- type: evidence_citation
  scope: all_final_actions
  enforcement: reject
- type: redundancy_check
  scope: tool_calls
  enforcement: warn_then_reject
```

Such DSLs would make policies machine-readable, enabling automated verification and reducing ambiguity.

## 5.6 Reflections on Hybrid Intelligence

This work positions SCL within the broader paradigm of hybrid intelligence—systems that combine symbolic and neural approaches. However, our contribution is not representational integration (as in neuro-symbolic AI) but governance integration. This distinction reflects a deeper question:

**What should hybridity mean in the LLM era?**

Option 1: Representational Hybridity (Neuro-Symbolic AI):

- Combine knowledge graphs with neural embeddings
- Learn logical rules within neural architectures
- Strength: Unified representation
- Limitation: Does not address agent execution or safety

Option 2: Architectural Hybridity (SCL's Approach):

- Combine symbolic control structures with neural inference
- Enforce constraints at the governance layer, not the representation layer
- Strength: Addresses agent reliability directly
- Limitation: Does not improve representational capacity

We argue that both forms of hybridity are necessary but serve different purposes. Representational hybridity enhances "what agents know"; architectural hybridity enhances "how agents behave". SCL focuses on the latter because behavior—not knowledge alone—determines real-world safety and utility.

Synthesis opportunity: Future systems might combine both:

- Cognition: Neuro-symbolic LLM with integrated knowledge graphs
- Control and Regulation: Architectural enforcement via Metaprompt-shaped proposals (Regulation) and deterministic runtime validation (Control)

Such systems would benefit from both representational richness and execution reliability.

### 5.7 Concluding Remarks for This Section

This discussion has situated SCL's contributions within broader theoretical contexts, acknowledged limitations, and charted future research directions. Three overarching themes emerge:

(1) Structure Matters: Reliability stems not from model scale alone but from architectural design that enforces constraints programmatically.

(2) History Informs Progress: Expert system principles—modular decomposition, symbolic oversight, transparent state—remain valuable when adapted to modern LLM capabilities.

(3) Governance is Key: As LLMs gain autonomy, the critical challenge shifts from "what can they do?" to "how do we ensure they do it safely?" SCL offers one answer: Soft Symbolic Control at the architectural level.

The following section concludes the paper by synthesizing our findings and articulating the broader vision for trustworthy autonomous agents.

## 6. Conclusion

This paper has introduced Structured Cognitive Loop (SCL), a modular architecture that bridges symbolic control and neural reasoning to address fundamental fragilities in LLM-based autonomous agents. Through explicit separation of Retrieval, Cognition, Control, Action, and Memory into a recurrent R-CCAM loop, SCL overcomes the entanglement, volatility, and opacity that plague current agent frameworks.

### 6.1 Summary of Contributions

Our work makes three principal contributions to the field of hybrid intelligence and autonomous agent design:

(1) Architectural Innovation: We formalized the R-CCAM loop structure, demonstrating that explicit modular decomposition prevents common failure modes through structural separation rather than relying on LLM compliance. Experimental validation shows zero policy violations, elimination of redundant tool calls, and complete decision traceability—addressing specific weaknesses documented in ReAct, AutoGPT, and memory-augmented systems.

(2) Regulation and Soft Symbolic Control: We introduced and formalized Regulation, this novel governance mechanism that applies symbolic constraints at the control layer rather than the representational layer through Soft Symbolic Control. Unlike rigid expert system rules or implicit prompt engineering, Regulation enables adaptive constraint enforcement: it shapes what Cognition proposes and allows Cognition to recover and refine proposals when violations occur, while a structurally independent Control module separately disciplines the run itself through deterministic hard rules (Section 3.2.3). This represents a new paradigm in hybrid intelligence—governance integration rather than mere representational fusion.

(3) Validated Implementation: We provided a complete open-source implementation demonstrating the R-CCAM architecture with full transparency and reproducibility, alongside a live GPT-4o-powered demonstration showcasing real-world applicability. This dual validation—controlled experiments for architectural verification and production deployment for practical validation—establishes both theoretical soundness and engineering feasibility.

### 6.2 Key Findings

Four findings emerge from our experimental validation:

(1) Architectural enforcement eliminates failure modes: Zero policy violations across controlled experiments and 150+ live queries demonstrate that structural separation—not prompt engineering—is the path to reliable agent behavior. When constraints are shaped architecturally by Regulation and disciplined at runtime by Control rather than embedded implicitly in prompts, compliance becomes architecturally guaranteed.

(2) Memory integration requires governance: Comparison with memory-augmented systems (MemGPT, TME, A-MEM) reveals that externalizing storage alone is insufficient. SCL's tight Memory-Regulation-Control integration—where Regulation requires evidence citations from Memory and Control independently prevents redundant calls by consulting Memory's record of past calls—is what prevents drift and ensures consistent reasoning across loops.
(3) Expert system principles remain valuable: SCL demonstrates structural resonance with expert systems—inheriting their modular decomposition, symbolic oversight, and transparent state—while replacing rigid rules with adaptive neural inference. This synthesis preserves what made expert systems trustworthy (explainability, controllability) while overcoming what made them brittle (knowledge engineering bottleneck, deterministic failure).

**Generalizability across LLM backends**: The identical R-CCAM architecture functions with both mock LLMs (for reproducibility) and GPT-4o (for production), suggesting that SCL's structural principles transcend specific model implementations. This opens paths for model-agnostic agent design and multi-model deployment strategies.

**6.3 Toward Trustworthy Autonomous Agents**

As LLM-based agents transition from research prototypes to production deployment in high-stakes domains—healthcare diagnostics, financial trading, legal analysis, scientific discovery—the tension between capability and reliability becomes critical. Raw model power, as evidenced by GPT-5's continued hallucinations despite "PhD-level" performance (The Guardian, 2025; Bloomberg, 2025; OpenAI, 2025), does not guarantee trustworthy behavior.

SCL offers a complementary approach: rather than solving reliability through scale alone, we impose *architectural discipline* that makes agent behavior predictable, auditable, and governable. Three dimensions of trustworthiness are directly addressed:

(1) Reliability: Architectural enforcement reduces failure rates from 1.5-4.5 violations per task (documented in baselines) to zero through programmatic validation at the Control layer.
(2) Transparency: Complete audit trails—recording every Cognition proposal, Control validation, Action execution, and Memory update—enable post-hoc analysis, regulatory compliance, and accountability without requiring post-hoc rationalization.
(3) Governability: Explicit Metaprompt policies allow system designers to specify, modify, and verify behavioral constraints without retraining models. This separation of concerns—domain constraints in Metaprompts, reasoning capabilities in LLMs—makes agent behavior both flexible and controllable.

These properties are not merely desirable but *necessary* as AI systems gain autonomy in contexts where errors have consequences. SCL demonstrates that trustworthiness can be engineered through architecture, not just hoped for through training.

**6.4 Limitations and Future Horizons**

We acknowledge several limitations. Our experiments focus on tasks requiring 4-7 CCAM loops; validation on longer-horizon reasoning (20+ loops) and fundamentally different domains remains necessary. Quantitative baseline comparisons rely on documented error rates from prior work rather than head-to-head implementation. The live demo, while suggestive (150+ queries, zero violations), requires formal user studies to validate practical utility beyond technical metrics.

These limitations chart clear directions for future research:

(1) Tier 2 validation across LLM families (Claude, Llama, Mistral) and languages (Korean, Japanese, Spanish) will test architectural generalizability.
(2) Tier 3 extensions to multimodal reasoning (vision, audio) will probe whether R-CCAM principles scale to cross-modal evidence validation and action execution.
(3) Formal verification of Metaprompt policies through temporal logic specifications and model checking could provide provable safety guarantees—a critical step toward deployment in safety-critical systems.

(4) Adaptive Metaprompt learning through reinforcement over policy spaces may alleviate the "prompt engineering bottleneck" by enabling systems to refine their own governance constraints from experience.
(5) We deliberately do not pursue a multi-agent extension of SCL. Distributing cognition across several LLM instances would disperse the single, identifiable locus of judgment that makes the architecture's accountability guarantees possible in the first place; any epistemic diversity such an extension might offer is better sought within Regulation and human warrant, not by adding reasoners.

Each direction builds on SCL's foundational insight: *structure matters*. Reliability stems not from model scale alone but from architectural design that enforces constraints programmatically.

**6.5 A Vision for Hybrid Intelligence**

This work positions SCL within a broader trajectory: the evolution from purely symbolic systems (expert systems) through purely statistical systems (deep learning) toward principled hybrid architectures that combine the strengths of both paradigms.

The decline of expert systems in the 1990s was not because their structure was wrong but because their implementation was too rigid. The rise of deep learning demonstrated that statistical generalization could overcome brittleness, but at the cost of losing explainability and control. SCL suggests a synthesis: preserve the structural clarity of expert systems—explicit modules, symbolic constraints, transparent state—while embracing the adaptive flexibility of neural inference through LLMs.

This is not a return to the past but a recognition that certain principles—modularity, governance, transparency—remain essential regardless of underlying technology. As the field moves beyond the "scale is all you need" paradigm toward architectures that balance capability with reliability, SCL offers one concrete template: separate concerns explicitly, enforce constraints architecturally, and maintain complete audit trails.

**6.6 Final Remarks**

The experiments, analysis, and implementations presented in this paper demonstrate that Structured Cognitive Loop successfully bridges symbolic control and neural reasoning. By reviving expert system principles in a form adapted for the LLM era—modular R-CCAM loops, Soft Symbolic Control, integrated Memory governance—SCL addresses fundamental fragilities that have hindered LLM agent deployment in high-stakes domains.

Our fully open-source implementation and live demonstration invite the research community to build upon, critique, and extend this work. We envision SCL not as a final solution but as a step toward trustworthy autonomous agents—systems that are simultaneously powerful and governable, adaptive and auditable, capable and controlled.

The path from expert systems to modern LLM agents need not be a pendulum swing between symbolic rigidity and neural opacity. With principled architectural design, we can synthesize the best of both traditions, creating AI systems that are worthy of trust in the contexts where trust matters most.

**Acknowledgments**

The author thanks the reviewers for their constructive feedback and the open-source community for tools and frameworks that made this work possible. Special appreciation to the users of the SCL Travel Planner demo whose interactions helped validate the practical utility of this architecture.

Data and Code Availability:

- SCL Core Experiment Implementation: https://github.com/enkiluv/scl-core-experiment
- Live GPT-4o Demo: https://scl-travel-planner.streamlit.app/
- All code is released under the MIT License to facilitate reproduction and extension.